\documentclass[10pt,twocolumn,letterpaper]{article}

\usepackage{iccv}
\usepackage{times}
\usepackage{epsfig}
\usepackage{graphicx}
\usepackage{amsmath}
\usepackage{amssymb}
\usepackage{float}
\restylefloat{table}
\graphicspath{{images/}}

\usepackage[pagebackref=true,breaklinks=true,letterpaper=true,colorlinks,bookmarks=false]{hyperref}

\iccvfinalcopy 

\def\iccvPaperID{23} 

\ificcvfinal\fi

\begin{document}

\title{Learning Unified Embedding for Apparel Recognition}

\author{Yang Song\\
Google\\
{\tt\small yangsong@google.com}
\and
Yuan Li\\
Google\\
{\tt\small liyu@google.com}
\and
Bo Wu\\
Google\\
{\tt\small bowu@google.com}
\and
Chao-Yeh Chen\\
Google\\
{\tt\small chaoyeh@google.com}
\and
Xiao Zhang\\
Google\\
{\tt\small andypassion@google.com}
\and
Hartwig Adam\\
Google\\
{\tt\small hadam@google.com}
}

\maketitle
\begin{abstract}
   
In apparel recognition, specialized models (e.g. models trained for a particular vertical like dresses) can significantly outperform general models (i.e. models that cover a wide range of verticals). Therefore, deep neural network models are often trained separately for different verticals (e.g. \cite{WTB2015}). However, using specialized models for different verticals is not scalable and expensive to deploy. This paper addresses the problem of learning one unified embedding model for multiple object verticals (e.g. all apparel classes) without sacrificing accuracy. The problem is tackled from two aspects: training data and training difficulty. On the training data aspect, we figure out that for a single model trained with triplet loss, there is an accuracy sweet spot in terms of how many verticals are trained together. To ease the training difficulty,  a novel learning scheme is proposed by using the output from specialized models as learning targets so that L2 loss can be used instead of triplet loss. This new loss makes the training easier and make it possible for more efficient use of the feature space.  The end result is 
a unified model which can achieve the same retrieval accuracy as a number of separate specialized models, while having the model complexity as one. The effectiveness of our approach is shown in experiments.
     
\end{abstract}
\section{Introduction}

 Apparel recognition has received increased attention in vision research (\cite{WTB2015,Yan2015,Liu_2016_CVPR,LLzisserman,LLsimo,LLserge}). Given a piece of garment, we want to find the same or similar items. This technology has great potential in assisting online shopping and improving both image search and mobile visual search experience.
 
 Apparel retrieval is a challenging problem. The difficulties are multifold. It is an object instance recognition problem. The appearance of the item changes with lighting, viewpoints, occlusion, and background conditions. For apparels, the images from online shopping sites may differ from those taken in ``real life" under uncontrolled conditions (also called street photos \cite{WTB2015}). Different verticals (or categories) also have different characteristics. For instance, images from the \textit{dress} vertical may undergo more deformations than those from the \textit{handbags} vertical.

\begin{figure}[t]
\begin{center}
\includegraphics[width=8cm]{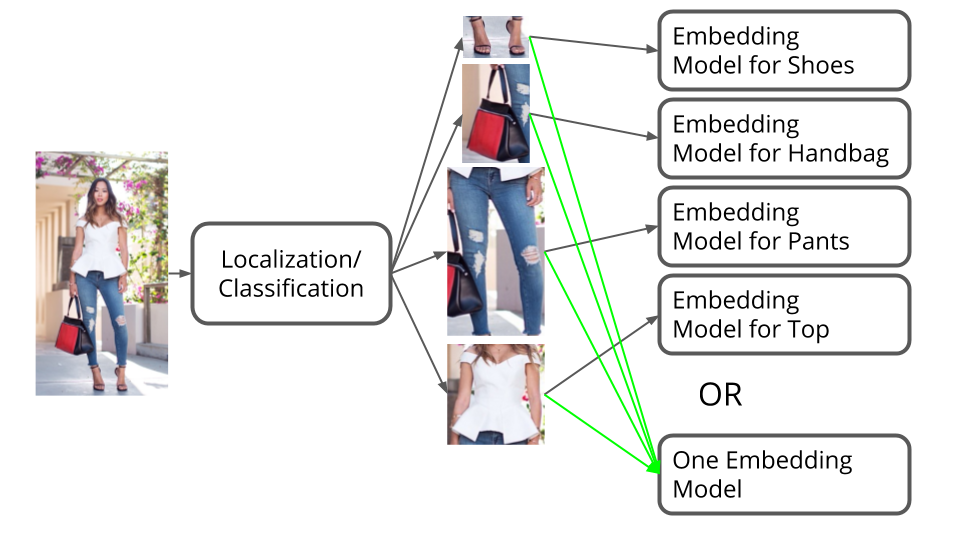}
\end{center}
   \caption{The paper addresses the following question: can a unified embedding model be learned across all the verticals in apparel recognition?}
\label{fig:overall}
\end{figure}
 
 In fine-grained/instance recognition, separate models are often used for different verticals. For example, in \cite{LinFGICCV2015,KrauseECCV2016}, separate models are built for birds, dogs, aircrafts, and cars. Similarly, in apparel recognition, separate models are trained for different verticals/domains (\cite{Yan2015,WTB2015}). In \cite{Yan2015}, the embedding models for images from shopping sites and from streets are learned using separate sub-networks. In \cite{WTB2015}, the network for each vertical (such as \textit{dress, handbags, sunglasses, and pants}) is fine-tuned independently in the final model training. While using separate models can help improve accuracy, it brings extra burden for model storage and deployment. The problem becomes more severe when the models are used on mobile devices. Therefore it is desirable to learn a unified model across different apparel verticals. 
 
 This paper studies the problem of learning unified models for apparel recognition. Our goal is to build a unified model which can achieve comparable retrieval accuracy as separate models, with the model complexity no bigger than a single specialized model. As shown in (Figure~\ref{fig:overall}), the clothing item is first detected and localized in the image. An embedding (a vector of floats) is then obtained from the cropped image to represent the item and is used to compare the similarity for retrieval. We focus on the embedding model learning in this paper. 
 
 One way to learn the unified model is to combine training data from different verticals. As shown in our experiments (Section ~\ref{sec:exp_combine_data}) and in \cite{WTB2015}, data combination may cause performance degradation. To avoid the performance degradation, we have developed a selective way to do vertical combination. Unfortunately, such ``smart'' data combination strategies are not enough - we cannot learn one unified model with satisfying accuracy. Is it possible to obtain such a model? Is the limitation intrinsic in model capacity or is it because of the difficulties in model training? Triplet loss is used to learn embedding for individual verticals, which has shown powerful results in embedding learning \cite{wang2014,SchroffKP15}. However, as noted in \cite{SohnNIPS16,SchroffKP15} and also observed in our experiments, triplet-based learning can be hard due to slow convergence and the nuances in negative sampling strategy. In this work, we seek new approaches to ease the difficulty in triplet training so that a unified model can be learned.

 This paper presents a novel way to learn unified embedding models for multiple verticals. There are two stages in model training. The first stage tackles a relatively easier problem - learning embedding models for individual verticals or a small number of combined verticals. In the second stage, the embeddings from the separate models are used as learning target and L2 loss is deployed to train the unified model. The second stage uses the feature mapping learned in the first stage, and combines them into one single model. As shown in Figure~\ref{fig:tsne} and Section~\ref{sec:combine_model}, the learned unified model can make better and broader use of the feature space.  
 
In summary, this paper proposes a two-stage approach to learn a unified model for apparel recognition. The new approach can help alleviate the training difficulty in triplet-based embedding learning, and it can make more efficient use of the feature space. We have also developed ways to combine data from different verticals to reduce the number of models in the first stage. As a result, a unified model is successful learned with comparable accuracy with separate models and with the same model complexity as one model.

The rest of the paper is organized as follows. Section~\ref{sec:individual} describes how feature embeddings are learned for individual verticals. Our work on how to learn a unified model across verticals is presented in Section~\ref{sec:unified}. Experiments are shown in Section~\ref{sec:experiments}, and it concludes at Section~\ref{sec:conclusion}.

\section{Learning Individual Embedding Models}\label{sec:individual}
As shown in Figure~\ref{fig:overall}, we adopt a two-step approach in extracting embedding feature vectors for object retrieval. The first step is to localize and classify the apparel item. Since the object class label is known from the first step, specialized embedding models can be used in the second step to compute the similarity feature for retrieval. We describe how we train the specialized embedding models in this section. Unified model learning will be depicted in section \ref{sec:unified}.

\subsection{Localization and Classification}\label{sec:localizer}
An Inception V2 (\cite{ioffe2015batch}) based SSD (\cite{LiuSSD15}) object detector is used. Other object detection architecture and base network combination can also work \cite{huang2016od}. This module provides bounding boxes and apparel class labels, i.e., whether it is a handbag or a pair of sunglasses or a dress. Features are then extracted on the cropped image using an embedding model.

\subsection{Embedding Training with triplet loss} \label{sec:triplet}
We use triplet ranking loss \cite{wang2014,SchroffKP15} to learn feature embeddings for each individual vertical. A triplet includes an anchor image, a positive image, and a negative image. The goal for triplet learning is to produce embeddings so that the positive image gets close to the anchor image while the negative is pushed away from the anchor image in the feature space. The embeddings learned from triplet training are suitable for computing image similarity. Let $t_i = (I_i^a, I_i^p, I_i^n)$ be a triplet, where $I_i^a, I_i^p, I_i^n$ represent the anchor image, positive image and negative image respectively. The learning goal is to minimize the following loss function, 
\begin{equation}\label{eq:hinge_loss}
  \begin{aligned}
  &l(I_i^a, I_i^p, I_i^n) = \\
   &\max \{0, \alpha + D(f(I_i^a), f(I_i^p)) - D(f(I_i^a), f(I_i^n)) \}
  \end{aligned}
\end{equation}
where $\alpha$ is the margin enforced between the positive and negative pairs, $f(I)$ is the feature embedding for image $I$, and $D(f_x, f_y)$ is the distance between the two feature embeddings $f_x$ and $f_y$.

In our applications, the positive image is always of the same product as the anchor image, and the negative image is of another product but in the same vertical. Semi-hard negative mining \cite{SchroffKP15} is used to pick good negative images online to make the training effective.

\subsubsection{Network Architecture}
\begin{figure}[htb]
\begin{center}
\includegraphics[width=8cm]{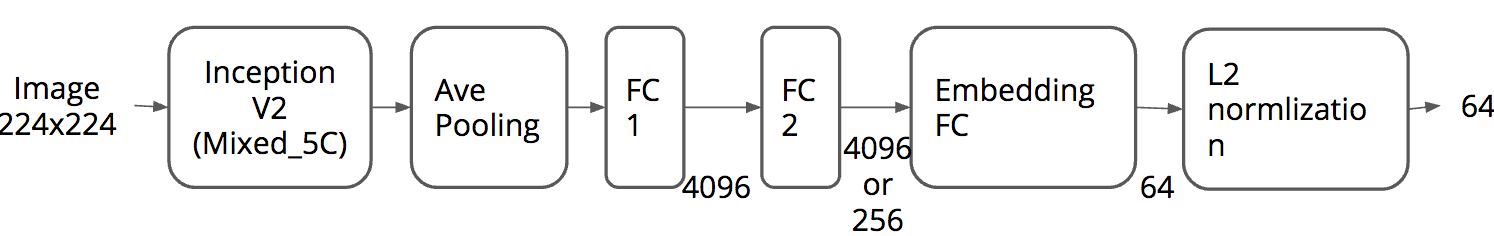}
\end{center}
   \caption{Network architecture for feature extraction. ``FC'' means fully connected layer, and the numbers are the output dimension of the layer.}
\label{fig:inceptionv2}
\end{figure}

Figure \ref{fig:inceptionv2} shows the network architecture. We use Inception V2 (\cite{ioffe2015batch}) as the base network, chosen mainly for efficiency reasons. Any other base network (e.g. \cite{He15ResNet,simonyan2014very,szegedy2016inception,szegedy2015rethinking}) can also be used.

\section{Learning Unified Embedding}\label{sec:unified}
 Section \ref{sec:individual} shows how embeddings for individual verticals are learned. Given enough training data for each vertical, good performance can be achieved. However, with more verticals in the horizon, having one model per vertical becomes infeasible in real applications. This section describes how a unified model across all verticals is learned. 
 
\subsection{Combining Training Data} \label{sec:combine_data}
One natural way to learn a model which can work for multiple verticals is to combine training data from those verticals. With the combined training data, models can be learned in the same way as described in Section \ref{sec:individual}. 

However as shown in our own experiments (Section \ref{sec:exp_combine_data}) and in \cite{WTB2015}, training models with combined data may cause accuracy degradation compared to models trained for each individual vertical. To prevent performance degradation, a greedy strategy is developed to decide data from which verticals can be combined. Starting from one vertical, we add data from other verticals in to see if the model learned from the combined data causes accuracy degradation. We keep adding until degradation is observed and keep the previous best combination of verticals. We end up with a number of specialized models, each covering a subset of verticals, while maintaining the best possible accuracy. In our experiments, this results in four specialized models for all apparel verticals.

\subsection{Combining Specialized Models} \label{sec:combine_model}
Combining the training data can only somewhat alleviate the coverage scalability issue. Is it possible to learn a unified model with the sample model complexity as one model and no accuracy degradation? Is model capacity the bottleneck or the difficulty in training?

Deep neural networks can be hard to train. The challenge of triplet training has been documented in literature \cite{wang2014,SchroffKP15,SohnNIPS16}. As exemplified in the Resnet work by He \etal in \cite{He15ResNet}, making the training easier can lead to substantial performance improvement. We propose a solution from a similar angle -- to ease the difficulty in model training.

We want to learn a unified model such that the embeddings generated from this model is the same as (or very close to) the embeddings generated from separated specialized models. Let $V=\{V_i\}_{i=1}^K$, where each $V_i$ is a set of verticals whose data can be combined to train an embedding model (Section \ref{sec:combine_data}). Let $M=\{M_i\}_{i=1}^K$ be a set of embedding models, where each $M_i$ is the model learned for vertical set $V_i$. Let $I=\{I_j\}_{j=1}^N$ be a set of $N$ training images. If the vertical-of- $I_j \in V_s$, $s=1 \dots K$, its corresponding model $M_s$ is used to generate embedding features for image $I_j$. Let $f_{sj}$ denote the feature embeddings generated from $M_s$ for image $I_j$. We want to learn a model ${U}$, such that the features produced from model $U$ are the same as features produced from separate models. Let $f_{uj}$ denote the feature embeddings generated from model $U$. The learning goal is to find a model $U$, which can minimize the following loss function,
\begin{equation}\label{eq:l2_loss}
  \begin{aligned}
  L = \sum_{j=1}^N \| f_{uj} - f_{sj} \|^2
  \end{aligned}
\end{equation}
Note that features $f_{uj}$ is computed from model $U$, while $f_{sj}$ may be computed from different models.

The above learning uses L2-loss, instead of triplet loss. L2-loss is easier to train than triplet loss. It is also easier to apply learning techniques such as batch normalization \cite{ioffe2015batch} on L2-loss.
The above approach allows the use of more unlabeled data. In triplet loss, the product identity (e.g. ``Chanel 2.55 classic flap bag'') is needed for generating the training triplet. Here only the vertical labels are needed, which can be generated automatically by the localization/classification model.

\subsubsection{Visualization}
\begin{figure}[htb]
  \begin{center}
  \includegraphics[width=8cm]{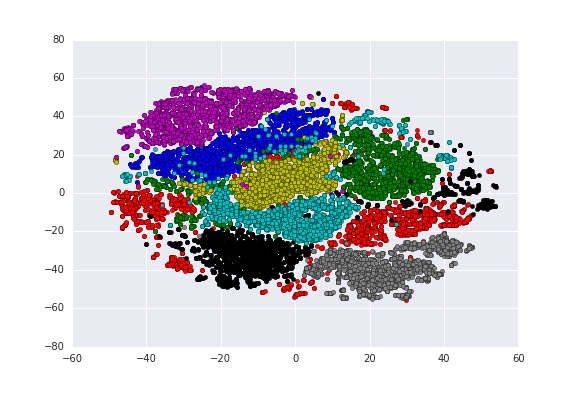}
  \end{center}
  \caption{The T-SNE projection for the embeddings. The original embedding are 64-d floats, 2000 samples from each apparel vertical. Blue: dresses; Red: footwear; Green: outerwear; Yellow: pants; Black: handbags; Grey: sunglasses; Cyan: tops;  Magenta: skirt.}
  \label{fig:tsne}
\end{figure}

The visualization of the features sheds lights on why our approach works. Figure \ref{fig:tsne} shows the  t-SNE projection (Barnes-Hut-SNE by Maaten \cite{tSne2013}) of the features generated from the separate models, i.e, $f_{sj}$. $f_{sj}$ is 64-d in our experiments. It includes two thousand images from each vertical, and the features are projected down to 2-d space for visualization. From Figure \ref{fig:tsne} we can see that the feature embeddings $f_{sj}$ are separated across verticals in the space. In other words, the embedding model for each vertical $f_{sj}$ (from model $M_s$) only uses part of the high dimensional (64-d in our case) space. Therefore one unified model can be learned to combine all of them. This answers our earlier question: the model capacity is not the bottleneck but rather the difficulty in training is.

\subsubsection{Relation to the Distillation work}
Our work is inspired by the distillation work in \cite{Hinton2015}. \cite{Hinton2015} focuses on classification models, and our work is to learn feature embeddings. In \cite{Hinton2015}, an ensemble of models are trained for the \textit{same} task, and then the knowledge in the ensemble is compressed into a single model. In contrast, the separate models $M_s$ in our work are trained for different tasks. As shown in Figure \ref{fig:tsne}, the feature embeddings from different verticals occupy different areas in the feature space. Our unified model is to consolidate multiple tasks in one model, and to make more efficient use of the feature space.
\section{Experiments} \label{sec:experiments}

\subsection{Training Data}\label{sec:training_data}
Our base network (Figure \ref{fig:inceptionv2}) is initialized from a model pre-trained using ImageNet (\cite{ILSVRC15}) data. For the triplet feature learning (Section \ref{sec:triplet}), there are two parts of training data. The training data are first collected from 200,000 search queries using Google Image Search.  A text query parser is run to get the apparel class (vertical) label for the text query. The text queries are from the following verticals, \textit{dresses, tops, footwear, handbags, eyewear, outerwear, skirts, and pants}. The search queries are specific product names crawled from online merchants. We take the top 30 images for each search query. The anchor and the positive images for the triplets are from the same search query, and the negatives are from a different search query, but in the same vertical as the anchor.  We call these triplets ``Image Search triplets''. We send a subset of triplets ($20,000$ triplets for each vertical) to human raters and verify the correctness of them. We call this second set of triplets ``clean triplets''.



For learning the unified embedding model (Section \ref{sec:combine_model}), we can use the same training data as for triplet feature learning. Since the unified embedding learning only needs the vertical label, which can be obtained via the localizer/classifier (Section \ref{sec:localizer}), it is possible to use more training data. However, in our experiments, we didn't observe significant performance improvement when using additional data. Therefore, in the unified model learning, the same training images are used as those in triplet embedding learning. 

\subsection{Evaluation Metrics}
The retrieval performance is measured by \textit{top-k accuracy}, i.e, the percentage of queries with at least one correct matching item within the first \textit{k} retrieval results. From the definition of the metric, for the same model, the bigger \textit{k} is, the higher the \textit{top-k accuracy} is. Data published in \cite{WTB2015} are used in evaluation. The unified model is also evaluated on \cite{Liu_2016_CVPR} data.


\subsection{Effect of Combining Training Data}  \label{sec:exp_combine_data}

\begin{table*}[htb]
\begin{center}
\begin{tabular}{|c|c|c|c|c|c|c|c|c|}
\hline
Method & bags & eyewear & footwear & dresses & tops & outerwear & pants & skirts \\
\hline
Individual models (no FT) & 57.5 & 53.1 & 26.6 & 48.6 & 24.8 & 26.7 & 25.5 & 37.1 \\
\hline
All data combined (no FT) & 55.6 & 35.8 & 25.1 & 30.9  & 18.5  & 17.4 &21.9&30.9 \\
\hline
Selected vertical combination (no FT) & 56.3 & 46.2 & 27.6 & 48.9 & 27.6 & 26.7 & 24.2 & 35.2\\
\hline
Selected vertical combination (with FT) & 66.9 & 48.3 & 35.7 & 59.1 & 35.2 & 29.6 & 27.6 & 46.4\\
\hline
\end{tabular}
\end{center}
\caption{ Comparison of top-1 retrieval accuracy. 
Individual Model = models trained individually on each vertical;  All data combined = the model trained with all verticals combined (one model);  Selected vertical combination = models trained with selected vertical combinations, four models in total (see text for details). ``FT'' means fine-tuning, indicating whether the models are fine-tuned with the clean triplets.}
\label{tab:vet_combine}
\end{table*}
This section presents our findings on combining different verticals of training data (Section \ref{sec:combine_data}). To calibrate the performance, triplet loss is first used to learn embeddings for each vertical (Section \ref{sec:triplet}). The goal for vertical combination is to use fewer number of models, but without retrieval accuracy degradation. 
The data from \textit{dresses, tops, footwear, handbags, eyewear, outerwear, skirts, and pants} verticals in \cite{WTB2015} are used in evaluation. These verticals are chosen because of their importance for our applications. The embedding models are based on the network depicted in Figure~\ref{fig:inceptionv2}, with the $FC 2$ output dimension being 4096-d.

The first three rows of Table~\ref{tab:vet_combine} show the \textit{top-1 accuracy}  of (1) models trained individually on each vertical; (2) the model trained with all verticals combined; (3) models trained with the following vertical combination. Training data from \textit{dresses and tops} are combined to train one model; \textit{footwear, handbags and eyewear} are combined to train one model; \textit{skirts and pants} are combined; \textit{outerwear} is trained on its own. This selected vertical combination is obtained by the method described in Section~\ref{sec:combine_data}. From Table~\ref{tab:vet_combine}, models trained with the selected vertical combination give comparable results with individual models. However,  the model trained with all verticals combined gives inferior results on some verticals such as eyewear, dresses, tops and outerwear.  This shows that it is not trivial to obtain a satisfying unified model by combining all the training data, and
we can only combine some verticals to achieve comparable accuracy with the individually trained models. 

The above models are trained using ``Image Search triplets''. To further improve the retrieval performance, ``Clean triplets'' are used to fine-tune the models. The last two rows of Table~\ref{tab:vet_combine} shows the \textit{top-1} accuracy comparison results. ``No FT'' models are trained with Image Search triplets, and the ``with FT'' models are further fine-tuned using clean triplets. This shows that fine-tuning with the clean data is an effective way to improve retrieval accuracy.

\subsection{Effect of Combining Models} \label{sec:exp_combine_model}
By combining the training data, the selected vertical combination results in four separate embedding models. A unified model for all the verticals is then learned via what we proposed in Section~\ref{sec:combine_model}. The unified embedding model is also based on the network depicted in Figure~\ref{fig:inceptionv2}, with the $FC 2$ output dimension being 256-d.  Therefore the unified model is even smaller in size than one single separated model. 

\begin{figure}[htb]
\begin{center}
\includegraphics[width=8cm]{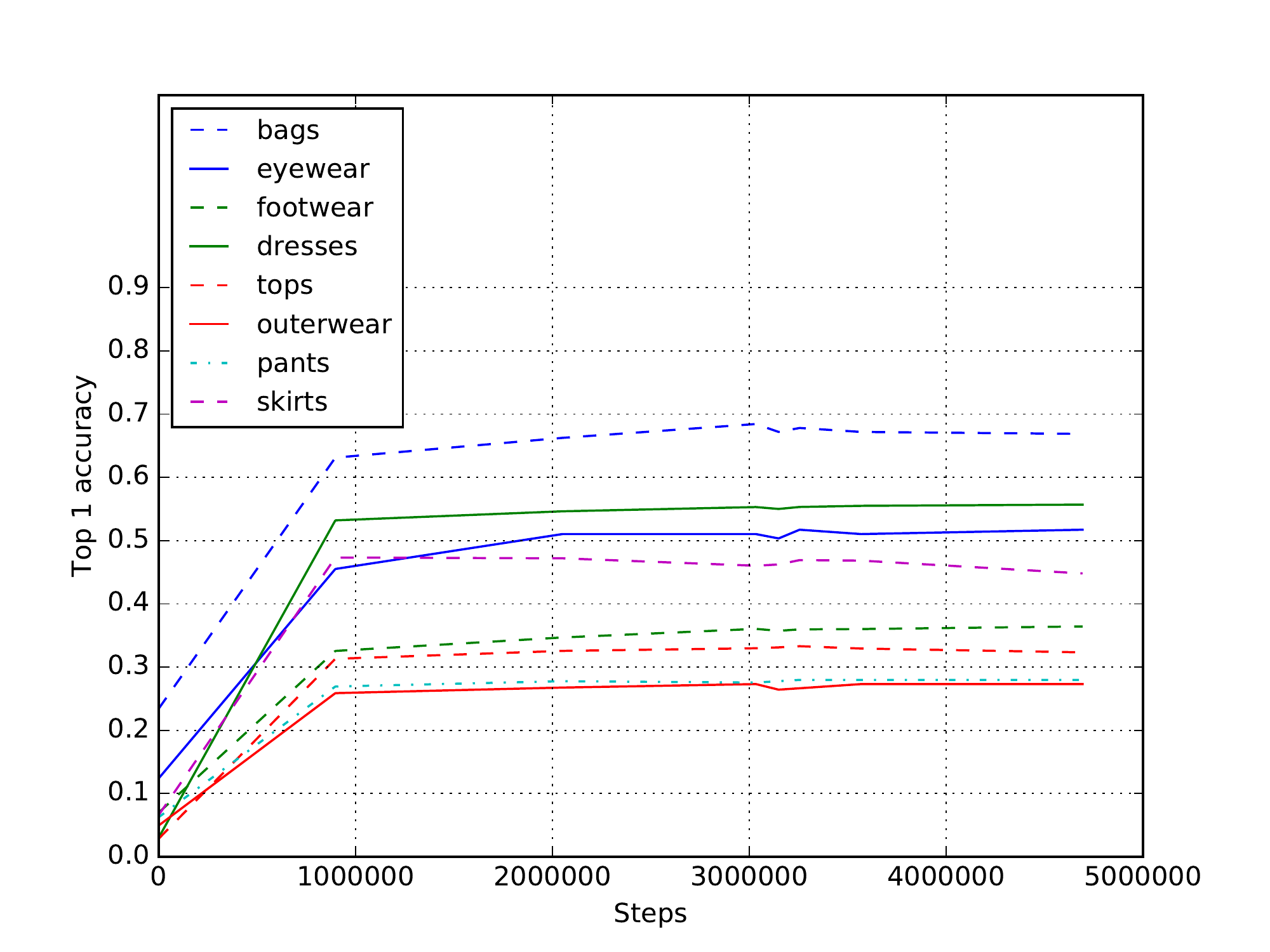}
\end{center}
   \caption{Top-1 accuracy vs. training steps.}
\label{fig:acu_steps}
\end{figure}

Figure~\ref{fig:acu_steps} shows how the \textit{top-1} accuracy changes with training steps (batch size is 32 for each step). Different verticals achieve highest accuracy at different steps. The model at step 3-million is chosen for further experiments.

\begin{table*}[htb]
\centering
\begin{tabular}{|c|c|c|c|c|c|c|c|c|}
\hline
Method & bags & eyewear & footwear & dresses & tops & outerwear & pants & skirts \\
\hline
WTB paper~\cite{WTB2015} (top-20) & 37.4 & 42.0  & 9.6 & 37.1 & 38.1 & 21.0 & 29.2 & 54.6\\
Unified Model (top-20)& \textbf{82.2} & \textbf{77.9} & \textbf{67.3} & \textbf{80.8} & \textbf{56.5} & \textbf{52.2} & \textbf{56.8} & \textbf{76.0} \\
\hline
Separate models (top-1) & 66.9 & 48.3 & 35.7 & \textbf{59.1} & \textbf{35.2} & \textbf{29.6} & 27.6 & \textbf{46.4} \\
Unified Model (top-1) & \textbf{68.4} & \textbf{51.0} & \textbf{36.0} & 55.4 & 33.0 & 27.3 & 27.6 & 46.0 \\
\hline
Separate models (top-5) & \textbf{76.3} & \textbf{64.1} & 52.4 & \textbf{74.6} & \textbf{49.2} & \textbf{45.9} & 43.2 & \textbf{62.4} \\
Unified Model (top-5) & 75.6 & 62.1 & \textbf{53.1} & 72.5 & 47.6 & 43.9 & \textbf{43.4} & 62.1 \\
\hline
\end{tabular}
\caption{Comparison of retrieval accuracy. The ``top-k'' inside the brackets shows which \textit{top-k} accuracy is evaluated. The ``Separate models" are trained with the selected vertical combination as in Section \ref{sec:exp_combine_data}. The ``Unified Model" is learned by the approach in Section \ref{sec:combine_model}.} 
\label{tab:results}
\end{table*}

Table~\ref{tab:results} shows the results of the unified model. The row "WTB paper" represents the best \textit{top-20} accuracy reported in paper~\cite{WTB2015} (Table 2 in that paper). The rows with ``Separate models'' are results from the selected vertical combination (Section~\ref{sec:exp_combine_data}). The rows with ``Unified Model'' are from the one unified model presented in this section using approach in Section~\ref{sec:combine_model}. Note that our models and the models from ~\cite{WTB2015} are trained from different training data.

\begin{figure}[htb]
\begin{center}
\includegraphics[width=8cm]{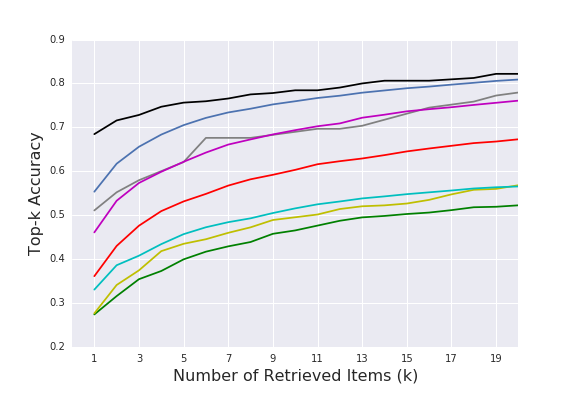}
\end{center}
   \caption{Top-k accuracy vs. the number of item retrieved (k). Black: handbags; Blue: dresses; Grey: eyewear; Magenta: skirts; Red: footwear; Cyan: tops; Yellow: pants; Green: outerwear.}
\label{fig:knn}
\end{figure}

The ``Separate models'' provide learning targets for the ``Unified Model''. In Table~\ref{tab:results}, the results from the ``Unified Model'' are very comparable to the results of ``Separate models''. For some verticals, the ``Unified Model'' performs even slightly better than ``Separate models''. We postulate two explanations. One is the natural variations in the evaluation metric; another reason is that the ``Unified Model'' can have better generalization performance than the ``Separate models'' since it is trained on data from all the verticals. Figure \ref{fig:knn} shows how the \textit{top-k} accuracy changes with the number of retrieved items (\textit{k}).

The unified model is also evaluated DeepFashion consumer-to-shop data \cite{Liu_2016_CVPR}. Using the ground-truth bounding boxes, our retrieval performance is 13.9\% (top-1) and 39.2\% (top-20), while it is 7.5\% (top-1) and 18.8\% (top-20)  in Fig 9(b) in \cite{Liu_2016_CVPR}. Note that the numbers are not  directly comparable as we use the GT bounding boxes. However, it serves the purpose of confirming the quality of our embedding model.

\begin{figure*}[htb]
\begin{center}
\includegraphics[width=12cm]{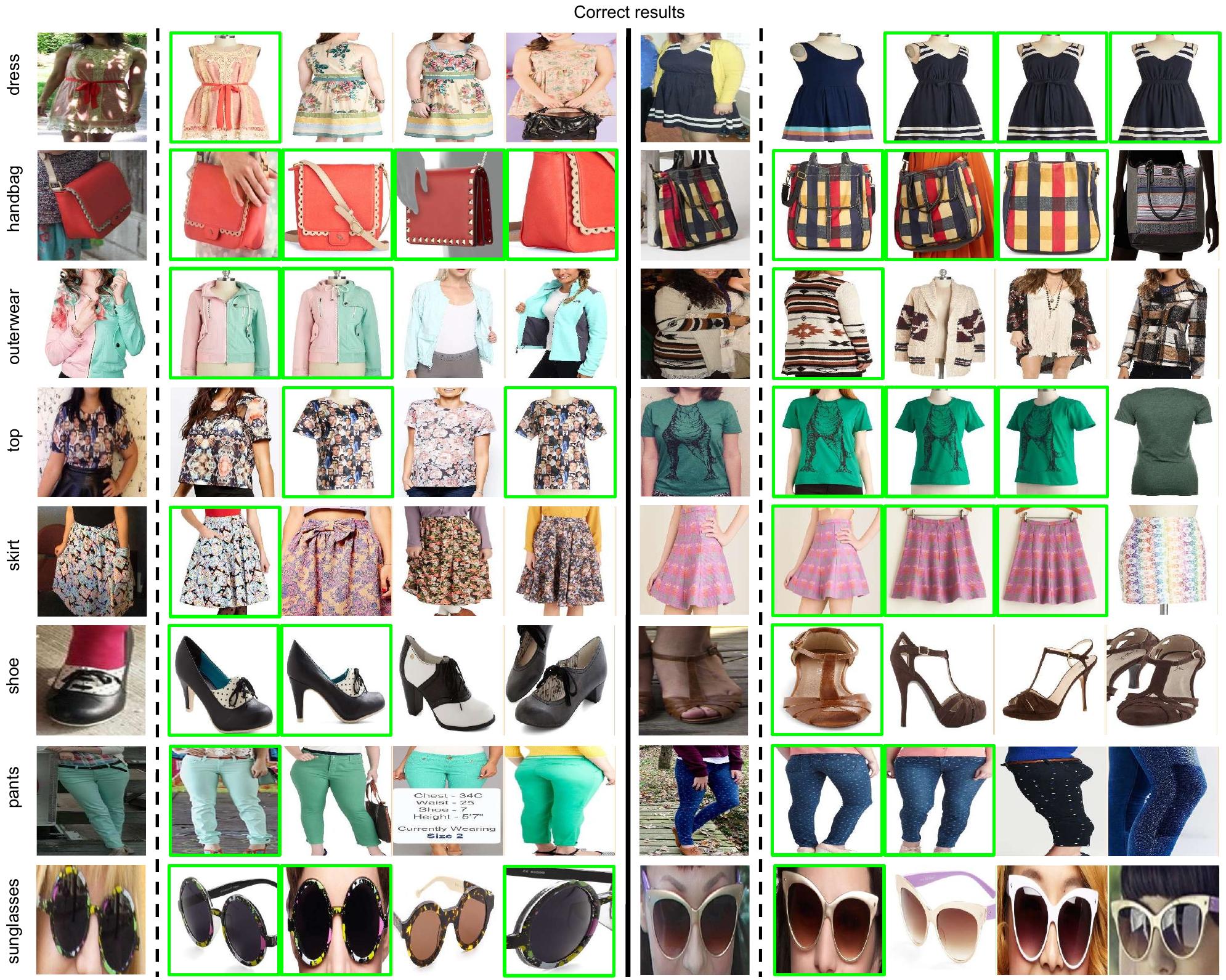}
\end{center}
   \caption{Retrieval results: successful examples. The images to left of the dashed lines are query images. The items in green bounding boxes are the correct retrieval results.}
\label{fig:good}
\end{figure*}
 
\begin{figure*}[htb]
\begin{center}
\includegraphics[width=12cm]{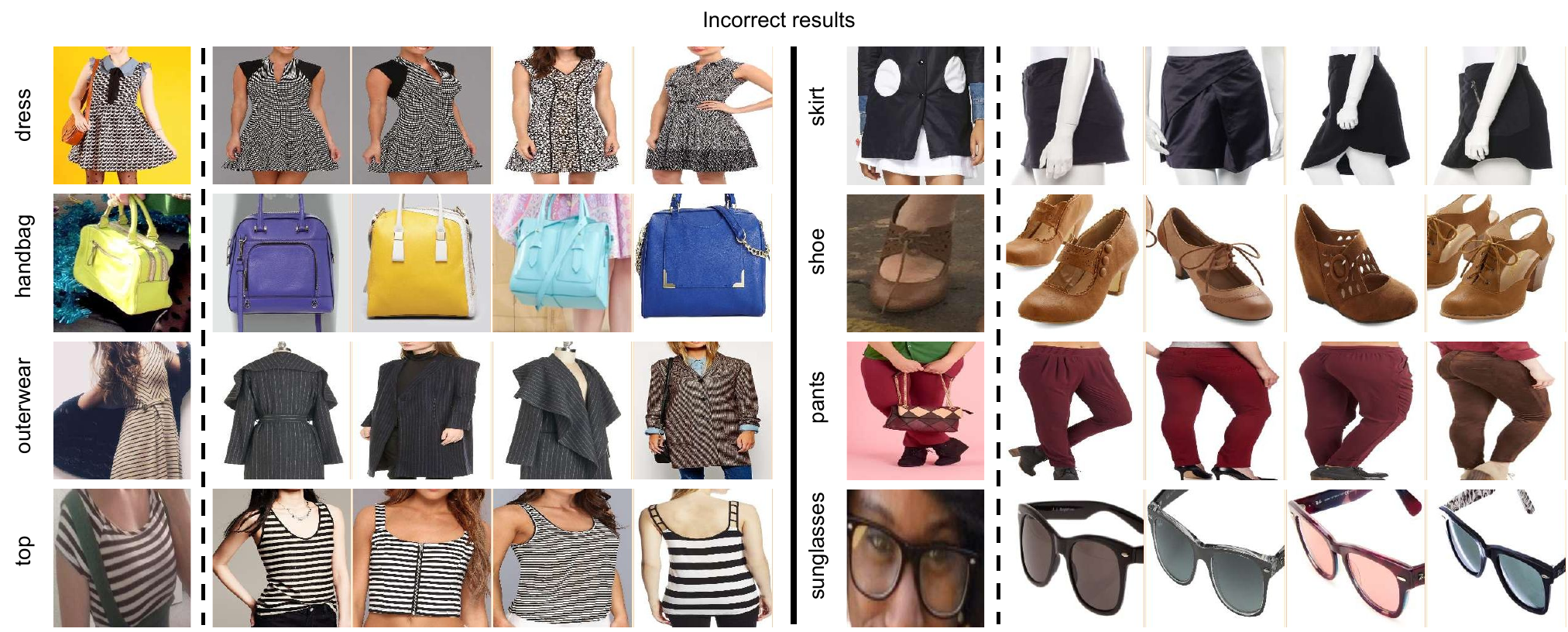}
\end{center}
   \caption{Retrieval results: not-so-successful examples (no correct item present in the top-4 returned results).}
\label{fig:bad}
\end{figure*}

\subsubsection{Retrieval Examples}
\vspace{-3mm}
Figures~\ref{fig:good} and \ref{fig:bad} show some sample retrieval results by using the ``Unified Model''. Figures~\ref{fig:good} shows the successful examples with correct item present in the top-4 returned results. Figures~\ref{fig:bad} gives some not-so-successful examples with no correct item present in the top-4 returned results. As shown in the figure, even when our model fails to get the correct product, the retrieved items are quite similar to the query. 

\section{Conclusion}\label{sec:conclusion}
This paper presents our approach and discoveries on how to learn a unified embedding models across all the apparel verticals. We figure out that for a single model trained with triplet loss, there is an accuracy sweet spot in terms of how many verticals are trained together. A novel way is proposed to ease the difficulty in training embeddings for multiple verticals. It uses embeddings from separate specialized models as learning target. The training becomes easier and makes full use of the embedding space. Successful retrieval results are shown on the learned unified model. The unified model has comparable accuracy with separate models and the same model complexity as one individual model. The unified model can make more efficient use of the feature space. Future work includes to extend this work to other fine-grained categories.

\bibliographystyle{ieee}
\bibliography{paper}

\end{document}